%% file: main.tex
\documentclass[sigconf]{acmart}
\usepackage{mismath}
\usepackage{multirow}
\usepackage[ruled,vlined, linesnumbered]{algorithm2e}
\usepackage{algpseudocode}
\usepackage[svgpath=tex/]{svg}
\usepackage{amsmath}
\usepackage{tikz}
\newcommand*\circled[1]{\tikz[baseline=(char.base)]{
            \node[shape=circle,draw,inner sep=2pt] (char) {#1};}}
\usepackage{tikz}
\usetikzlibrary{arrows}
\usepackage{tikz-network}

\AtBeginDocument{%
  \providecommand\BibTeX{{%
    \normalfont B\kern-0.5em{\scshape i\kern-0.25em b}\kern-0.8em\TeX}}}

\copyrightyear{2022}
\acmYear{2022}

\acmConference[IJCKG '22]{the 11th International Joint Conference on Knowledge Graphs}{October 27-28, 2022}{Hangzhou, China}

\begin{document}

\title{\textbf{RAILD}: Towards Leveraging Relation Features for Inductive Link Prediction In Knowledge Graphs}

\author{Genet Asefa Gesese}
\email{genet-asefa.gesese@fiz-karlsruhe.de}

\affiliation{%
  \institution{FIZ Karlsruhe -- Leibniz Institute for Information Infrastructure}
   \country{Germany}
}
\affiliation{
  \institution{Karlsruhe Institute of Technology, Institute AIFB }
  \country{Germany}
}
\author{Harald Sack}
\email{harald.sack@fiz-karlsruhe.de}

\affiliation{%
  \institution{FIZ Karlsruhe -- Leibniz Institute for Information Infrastructure}
   \country{Germany}
}
\affiliation{
  \institution{Karlsruhe Institute of Technology, Institute AIFB }
  \country{Germany}
}
\author{Mehwish Alam}
\email{mehwish.alam@fiz-karlsruhe.de}

\affiliation{%
  \institution{FIZ Karlsruhe -- Leibniz Institute for Information Infrastructure}
   \country{Germany}
}
\affiliation{
  \institution{Karlsruhe Institute of Technology, Institute AIFB }
  \country{Germany}
}

\renewcommand{\shortauthors}{F. Author, et al.}

\begin{abstract}
Due to the open world assumption, Knowledge Graphs (KGs) are never complete. In order to address this issue, various Link Prediction (LP) methods are proposed so far. Some of these methods are inductive LP models which are capable of learning representations for entities not seen during training. However, to the best of our knowledge, none of the existing inductive LP models focus on learning representations for unseen relations. In this work, a novel Relation Aware Inductive Link preDiction (RAILD) is proposed for KG completion which learns representations for both unseen entities and unseen relations. In addition to leveraging textual literals associated with both entities and relations by employing language models, RAILD also introduces a novel graph-based approach to generate features for relations. 
Experiments are conducted with different existing and newly created challenging benchmark datasets and the results indicate that RAILD leads to performance improvement over the state-of-the-art models.
Moreover, since there are no existing inductive LP models which learn representations for unseen relations, we have created our own baselines and the results obtained with RAILD also outperform these baselines. 
\end{abstract}

\ccsdesc[500]{Computing methodologies~Artificial intelligence}
\ccsdesc[500]{Computing methodologies~Machine learning}

\keywords{Knowledge graphs, Inductive link prediction, Textual descriptions, Entity representations, Relation representations}

\maketitle

\input{tex/Introduction}

\input{tex/Related-work}

\input{tex/Methodology}

\input{tex/Experiments}

\input{tex/Conclusion}


\bibliographystyle{ACM-Reference-Format}
\bibliography{main}



\end{document}

%% file: tex/Introduction.tex
\section{Introduction}

\label{sec:introduction}

Recently, Knowledge Graphs (KGs) have gained massive attention for use in various applications such as question answering, information retrieval, recommender systems, etc \cite{survey2017Wang}. Due to the open-world assumption
KGs are never complete~\cite{KnowVault2014} and hence, there arises the need for automated KG Completion (KGC) systems. In order to tackle this problem, various works~\cite{Ji2021Survey, gesese2019survey} have been done so far which are either rule-based techniques or distributed representation (embedding) learning methods. Many of the KGC approaches which are based on representation learning techniques utilize the well known Link Prediction (LP) task, i.e., the task of predicting missing links in the KG. The benefit of using an embedding-based LP task over rule-based approaches is that the embeddings of entities and relations learned in the LP task could also be leveraged in other downstream tasks.  

There are two major types of setups in LP tasks, i.e., transductive and inductive. In a transductive setup, all entities in the train and validation sets are required to be part of the training set. On the other hand, in an inductive setup, the validation and test sets may contain entities that are not seen during training. Even though most of the well known LP approaches~\cite{bordes2013translating,yang2015Distmult,Trouillon2016ComplEx,Seyed2018SimplE, sun2018rotate} are proposed for transductive settings, there are also several approaches~\cite{Daza2021InductiveER,Clouatre2021MLMLM} introduced that work in inductive settings. However, these representation-learning based inductive LP approaches do not pay attention to relations. Unlike entities for which there are textual descriptions that could be used as features for the entities, relations are usually just randomly initialized like in BLP~\cite{Daza2021InductiveER}. QBLP~\cite{Ali2021ImprovingIL} is a method that is proposed for inductive LP in hyper-relational graphs. This work could be generalized for unseen relations but since it does not provide a way to generate features for unseen relations, it could only be applied for inductive LP with relations involved during training. 

The most straightforward way to find features for relations is to use the 
descriptions of the relations. However, the issue  
is that the textual description could be either too short or entirely unavailable. In such cases, it is required to generate features for relations utilizing the structural information available at hand. This indicates that there is a need for a method which generates features for relations so that inductive LP could be performed with unseen relations. 

To this end, in this work, a novel approach \emph{Relation Aware Inductive Link preDiction (RAILD)} which predicts missing links in KGs considering both unseen entities and unseen relations is introduced. To the best of our knowledge, \emph{RAILD} is the first approach that handles unseen relations. It works by fine-tuning a pre-trained Language Model (LM) to encode textual descriptions of entities and relations. Moreover, it generates a graph-based relation features by first applying a novel algorithm named \emph{Weighted and Directed Network of Relations (WeiDNeR)} to build a directed relation-relation network from the triples available in the KG and then, generating embeddings for the relations in the network using Node2Vec model which leverages contextual information based on random walks. Then, these embeddings are in turn used as features for the relations for the LP task. Moreover, RAILD also utilizes the textual descriptions of the relations as features, by either combining them with the features generated by the feature generator component or separately.

\begin{figure}[htbp]
  \centering
  \includegraphics[width=250pt]{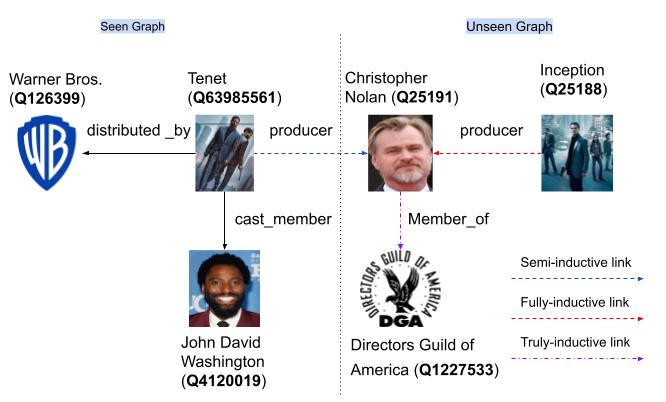}
  \caption{An example illustrating different settings of inductive LP tasks, i.e., semi-inductive (the link from \textit{Tenet} to \textit{Christopher Nolan}), fully-inductive (the link from \textit{Inception} to \textit{Christopher Nolan}), and truly-inductive (the link from \textit{Christopher Nolan} to \textit{Directors Guild of America}) settings}\label{fig:ilp}
\end{figure}

Figure~\ref{fig:ilp} provides an example 
of inductive LP settings followed in this work. Generally, inductive LP is divided into two categories: i) semi-inductive and ii) fully-inductive. In the semi-inductive setting, either the head or the tail is unseen but not both, and in the fully-inductive setting, both head and tail entities are unseen. In this definition, relations are often overlooked, i.e., they are usually assumed to be seen in the training set and hence are randomly initialized or as in MLMLM~\cite{Clouatre2021MLMLM}, they could be encoded using their labels (corresponding text descriptions) but without learning representations (embeddings) for them. Hence, this work divides the settings into three categories for clarity, i.e., semi-inductive (with seen relations), fully-inductive (with seen relations), and truly-inductive (with unseen entities and unseen relations).

This work formulates and addresses the following research questions: \textbf{Does encoding relations by generating features result in any performance improvement over State-Of-The-Art (SOTA) inductive LP models and also enable LP with unseen relations?} Following are the main contributions that are achieved in this work while attempting to answer the question: 
\begin{itemize}
    \item A novel algorithm is introduced to build a \textit{relation-relation network}, i.e., \textbf{WeiDNeR}, for the purpose of generating features for relations in a KG solely from the 
    contextual information present in the graph structure.  
    \item The results indicate that instead of randomly initializing relations in inductive LP, encoding the relations like the way entities are encoded by utilizing proper features leads to outperforming the SOTA inductive LP models on triple-based KGs.
    \item An experiment-supported evidence is provided showing that the algorithm introduced to generate features, WeiDNeR, enables producing competitive results as compared to using textual descriptions of relations as features.
    \item As part of the work, a novel LP dataset named \textbf{Wikidata68K}\footnote{https://doi.org/10.5281/zenodo.7066504} which contains unseen relations in the validation and test sets is introduced along with an automated pipeline to generate such datasets. Creating this dataset was required as there exists no such kind of evaluation dataset due to the fact that, to the best of our knowledge, no existing LP work deals with unseen relations. The results obtained with the proposed model on this challenging dataset are provided which could be seen as a first attempt to facilitate further research in the community on the topic of LP with unseen relations.      
\end{itemize}

This paper is organized as follows: Section~\ref{sec:related-work} discusses the related works in the area of inductive LP. In Section~\ref{sec:methodology}, the proposed model is presented followed by discussions on experimental results in Section~\ref{sec:experiments}. Section~\ref{sec:conclusion} concludes the work with some future directions.

%% file: tex/Related-work.tex
\section{Background and Related Works}\label{sec:related-work}

\subsection{Inductive LP Settings in triple-based KGs}
Existing inductive LP models operate 
either in \texttt{semi-inductive} setting where one of the head or tail entities is seen during training or \texttt{fully-inductive} setting where both head and tail entities are unseen during training. In both these settings, relations are usually assumed to be known during training. For the sake of clarity, in this work, predicting with unseen relations is defined separately named as \texttt{truly-inductive} LP setting. The formal definitions of these three settings are provided as follows:

Given a KG $G = (E,R)$ where $E$ and $R$ represent set of entities and relations respectively, let $T_{tr}$, $T_{va}$, and $T_{te}$ be sets of training, validation, and test triples where $E_{tr}$ \& $R_{tr}$, $E_{va}$ \& $R_{va}$, and $E_{te}$ \& $R_{te}$ are their corresponding set of entities and relations respectively. 
\paragraph{In \textbf{semi-inductive} setting} For every triple $<h,r,t> \in T_{va}$ or $<h,r,t> \in  T_{te}$, either or both of $h \in E_{tr}$ \& $t \in T_{tr}$ holds true while $R_{va} \subseteq R_{tr} $ and $R_{te} \subseteq R_{tr}$.

\paragraph{In \textbf{fully-inductive} setting} For every triple $<h,r,t> \in T_{va}$ or $<h,r,t>  \in T_{te}$, both $h \notin E_{tr}$ \& $t \notin E_{tr}$ holds true while $R_{va} \subseteq R_{tr} $ and $R_{te} \subseteq R_{tr}$.

\paragraph{In \textbf{truly-inductive} setting} For every triple $<h,r,t> \in T_{va}$ or $<h,r,t> \in  T_{te}$, either or both of $h \notin E_{tr}$ \& $t \notin E_{tr}$ holds true while there exist a set $R_v \subseteq R_{va}$ and a set $R_{t} \subseteq R_{te}$ where $R_{v} \nsubseteq R_{tr}$ and $R_{t} \nsubseteq R_{tr}$.

\subsection{Inductive LP Approaches}
Adapting most of the existing transductive LP models such as RotatE~\cite{sun2018rotate}, Distmult~\cite{yang2015Distmult}, ComplEx~\cite{Trouillon2016ComplEx}, and TransE~\cite{bordes2013translating} for inductive settings requires expensive re-training in order to learn embeddings for unseen entities. Therefore, such models are not applicable to making predictions with unseen entities. 
This led to the creation of some inductive LP approaches which are presented as follows.

\paragraph{\textbf{Rule-based methods}} Statistical rule-mining approaches make use of logical formulas to learn patterns present in KGs~\cite{Meilicke2018FineGrainedEO}. Despite the fact that such approaches are inherently applicable to inductive settings, they are prone to limited expressiveness and scalability issues. In order to address this issue, NeuralLP~\cite{yang2017NeuralLP} is proposed and it works by learning first-order logical rules in an end-to-end differentiable model. DRUM~\cite{Sadeghian2019DRUM} is another method that applies a differentiable approach for mining first-order logical rules from KGs and provides improvement over NeuralLP.
\paragraph{\textbf{Embedding-based methods}} Training entity encoders through feed-forward and graph neural networks is one way to generate representations for unseen entities as in GraphSAGE~\cite{Hamilton2017GraphSage}. However, such approaches require fixing a set of attributes (e.g., bag-of-words) before training in order to learn entity representations which leads to restricting their application on downstream tasks as discussed in~\cite{Daza2021InductiveER}.  
Aggregating neighborhood information through a graph neural network is one way to generate embeddings for entities~\cite{Hamaguchi2017transfer, Wang2019LAN}. The drawbacks of these approaches lie in the fact that they require the new nodes (i.e., the unseen entities) to be surrounded by known nodes and fail to handle entirely new graphs as discussed in~\cite{Teru2020GraIL}. KEPLER~\cite{Wang2021KEPLERAU} is a unified model for knowledge embedding (KE) and pre-trained language representation by encoding textual entity descriptions with a pre-trained LM as their embeddings, and then jointly optimizing the KE and LM objectives. However, due to the additional language modeling objective, KEPLER is quite expensive to compute and requires more training data. BLP~\cite{Daza2021InductiveER} utilizes a pretrained LM for learning representations of entities via a LP objective which is inspired by the work DKRL~\cite{Xie2016DKRL}. It demonstrates the power of LMs in facilitating the strong generalizability of entity embeddings on downstream tasks. QBLP~\cite{Ali2021ImprovingIL} is a model proposed to extend BLP for hyper-relational KGs by exploiting the semantics present in qualifiers. 

\paragraph{\textbf{Other Approaches}} 
GraIL~\cite{Teru2020GraIL} is a method that reasons over local subgraph structures to predict missing links in KGs. MLMLM~\cite{Clouatre2021MLMLM} proposes a mean likelihood method to compare the likelihood of different text of different token lengths sampled from a Masked LM to perform LP. Even though both of these approaches could predict missing links with unseen entities, they learn embeddings neither for entities nor for relations.

Note that the models discussed so far do not consider unseen relations except MLMLM which also does not learn embeddings at all. To address both limitations, i.e., considering unseen relations and also learning embeddings while predicting missing links, a novel method \texttt{RAILD} is proposed in this work for inductive LP in triple-based KGs. RAILD improves the SOTA methods such as BLP by introducing a separate encoder that utilizes structured information from KGs to generate features for relations. Moreover, in addition to the textual descriptions of entities, it also leverages the semantics present in the textual descriptions of relations by using a BERT encoder. Note that, in order to perform the LP task, both the structure-based and text-based encoders are combined.

%% file: tex/Methodology.tex
\section{RAILD: Relation Aware Inductive Link Prediction}\label{sec:methodology}

\begin{figure*}
    \centering
    \includegraphics[width=0.75\textwidth]{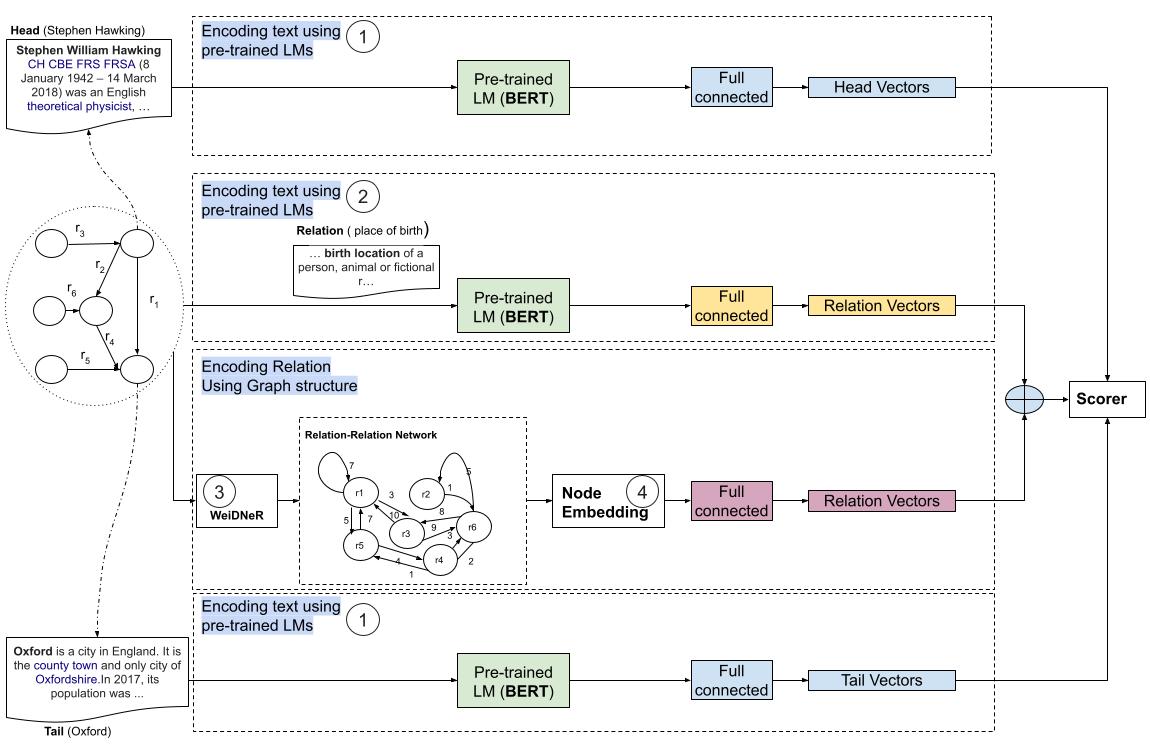}
    \caption{RAILD framework}
    \label{fig:framework}
\end{figure*}

The general architecture of the proposed approach is given in Figure~\ref{fig:framework}. As mentioned before, RAILD fine-tunes BERT pre-trained model 
to encode entities with an LP task.
Differently from BLP where relations are randomly initialized, in RAILD the same pre-trained BERT model is also applied to encode relations using their corresponding textual descriptions, as shown in Figure~\ref{fig:framework} component \circled{$2$}.  In addition to encoding relations using BERT, a feature generator component that is based solely on graph structure is also proposed. Hence, two kinds of vectors could be generated as features for relations, i.e., text-based and graph-based. Given a triple $<h,r,t>$, the two feature vectors generated for the relation $r$ are concatenated into a single vector. Since concatenation of the two relation vectors leads to doubling the output vector dimension, the vectors of the head/tail entities are also duplicated (concatenating the head vector with itself to match the size of the concatenated relation vector). Then, the resulting head $\textbf{h}$, tail $\textbf{t}$ and relation $\textbf{r}$ vectors are passed to the LP scoring function. 

Textual description encoding is performed by passing the text as input to a pre-retrained LM (specifically BERT but any other transformer based LM model could be used as well) and then passing the obtained vector from BERT through a feed-forward layer, as shown in Figure~\ref{fig:framework} component \circled{$1$} and \circled{$2$}. For the graph-based feature generation for relations, two major steps are applied, i.e., building a relation-relation network (Figure~\ref{fig:framework} component \circled{$3$}) and generating node embeddings for the created network where the nodes are relations (Figure~\ref{fig:framework} component \circled{$4$}). In the subsequent sections, the different components of the proposed model are presented in detail. First, encoding textual descriptions using pre-trained BERT is discussed followed by the description of the WeiDNeR algorithm. Then, the node embedding model applied in this work is presented. Finally, the chosen scoring functions are analyzed. 

\vspace{-1mm}
\subsection{Encoding Textual Descriptions using BERT}
Textual descriptions of entities contain information that would provide useful semantics while learning KG representations. In order to make use of such text data for representation learning, both static embedding models such as SkipGram~\cite{Mikolov2013SkipGram} and contextual embedding models like BERT have been extensively applied with different machine learning and natural language processing tasks. The power of transformer networks \cite{NIPS2017Attention} in encoding text to contextualized vectors has been well received. In particular, pre-trained embedding models such as BERT provide an advantage to fine-tune the model on other downstream tasks. 

In BLP, pre-trained BERT 
is fine-tuned on the inductive LP task and it showed promising results as compared to other methods. 
Our approach follows the same step, however, the difference is that i) in BLP the encoder is used only for entities whereas in our approach it is used also to encode relations, and ii) the fine-tuning is extended by adding an additional component with structure-based features for relations as shown in Figure~\ref{fig:framework}. 

Let $d=(w_1, ..., w_k)$ be an entity or relation description, the BERT tokenizer first adds two special tokens [CLS] and [SEP] to the beginning and end of $d$, respectively ($[CLS],w_1, . . . ,w_k, [SEP]$). BERT takes this as an input leading to a sequence of k + 2 contextualized embeddings as an output, i.e., $BERT(D) = [h_{CLS},h_1, . . . ,w_k, h_{SEP}]$. 

 

As in BLP and many other works which employ BERT for text encoding, this work also utilizes the contextualized vector $h_{CLS} \in \mathbb{R}$ where $h$ is the hidden size in the BERT architecture. Once $h_{CLS}$ is obtained, it will be given as an input to a linear layer that reduces the dimension of the representation, to yield the output entity or relation embedding $h = Wh_{CLS}$, where $w \in \mathbb{R}^{d \times h}$ is the weight with $d$ being the chosen embedding dimension. Note that as is shown in Figure~\ref{fig:framework}, the weights are shared with also the linear layer applied to the relation embeddings obtained with Node2Vec model.

\subsection{Weighted and Directed Network of Relations (WeiDNeR)}\label{subsec:WeiDNeR}

WeiDNeR is designed based on the following assumption. 
Given a KG $G=(R,E, T\subseteq E\times R\times E)$, $r_1, r_2 \in R$ and $T_1 \subseteq (T\cap (E\times {r_1}\times E))$, $T_2 \subseteq (T\cap (E\times {r_2}\times E))$, the assumption would be that the higher the number of common entities between $T_1$ and $T_2$, the higher the probability that $r_1$ and $r_2$ could be semantically similar.

Hence, based on this assumption, an algorithm is proposed which generates a directed and weighted network graph $N_{rel}=(V,L\subseteq V\times V, w)$ where the nodes $V$ are relations in the input KG (i.e., $V \subseteq R$), $L$ is the set of edges connecting the nodes, and $w: L\mapsto \mathbb{R}$ assigns weight to each edge. Algorithm~\ref{alg:WeiDNeR}, step by step, explains the process of creating the network graph. 

Following the procedure in this algorithm, if there is a direct link between two nodes \circled{$r_1$} and \circled{$r_2$} in the generated network $N_{rel}$ then the following statement holds true.

 $\big| head(T_1) \cap head(T_2) \big| > 0$  OR $\big| tail(T_1) \cap tail(T_2) \big| > 0$ OR $\big| tail(T_1) \cap head(T_2) > 0 \big|$

where $head(T_i)$ and $tail(T_i)$ are the sets of entities occurring at the head and tail positions in the set of triples $T_i$ respectively.  

If there is no direct link, then the statement becomes false, i.e., the two relations are not associated with any common entity in the input KG.

\noindent{\textbf{Algorithm description.}} Taking a KG $G=(R,E, T\subseteq E\times R\times E)$ with $\{< h_i, r_j, t_k > | < h_i, r_j, t_k > \in T \}$ where $h_i,t_k\in E$ and $r_j\in R$ as an input and generates a relation-relation network $N_{rel}$. For each pair of distinct relations ($r_a$, $r_b \in R$) it performs the following steps: 
\begin{itemize}
    \item it counts the number of pair of triples where the relation in the first triple is $r_a$ and in the second is $r_b$ and the tail entity in the first triple is the same as the head entity in the second triple (i.e., refer to line ~\ref{alg:dir1})..    
    
    \item it counts the number of pairs of triples where the relation in the first triple is $r_a$ and in the second is $r_b$ and the head entity in the first triple is the same as the tail entity in the second triple (i.e., refer to line ~\ref{alg:dir2}).
    
    \item it computes the number of entities shared by the triples associated with $r_a$ and $r_b$ at the exact same position at the head or at the tail, (i.e., refer to line ~\ref{alg:indir1}).
    
    \item If $\#direct + \#indirect > 0 $, then an edge from node $r_a$ to node $r_b$ will be created with the summed result given as a weight for the edge (i.e., refer to lines ~\ref{alg:sum1-starts} to ~\ref{alg:sum1-ends} ).
    
\end{itemize}
On the other hand, if $r_a$ and $r_b$ are the same, then the following will be performed.
\begin{itemize}
    \item  it counts the number of pairs of triples where the relations in both triples is $r_a$ and the tail entity in the first triple is the same as the head entity in the second triple (i.e., refer to line ~\ref{alg:dir3})   
    
    \item it computes the number of entities shared by the triples associated with $r_a$ at the exact same position at the head or at the tail. (refer to line ~\ref{alg:indir2})
    
    \item If $\#direct + \#indirect > 0 $, then an edge from node $r_a$ to node $r_b$ will be created with the summed result given as a weight for the edge (refer to lines ~\ref{alg:sum2-starts} to ~\ref{alg:sum2-ends} ).
\end{itemize}

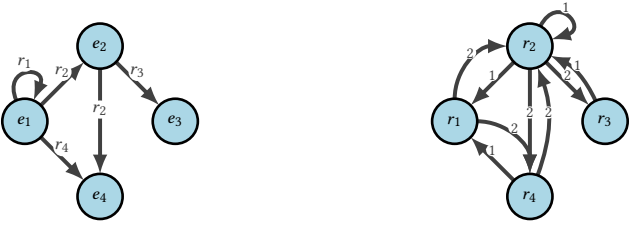
\begin{figure}
    \centering
\begin{tikzpicture}
\Vertex[label=$e_1$]{1}
\Vertex[x=1,y=1, label=$e_2$]{2}
\Vertex[x=2, label=$e_3$]{3}
\Vertex[x=1, y=-1, label=$e_4$]{4}
\Edge[Direct,label=$r_1$,position=above,loopsize=0.5cm, loopposition=90,loopshape=45](1)(1)
\Edge[Direct,label=$r_2$,position=above](1)(2)
\Edge[Direct,label=$r_4$,position=above](1)(4)
\Edge[Direct,label=$r_3$,position=above](2)(3)
\Edge[Direct,label=$r_2$,position=above](2)(4)

\end{tikzpicture}
\hspace{3cm}
\begin{tikzpicture}
\Vertex[x=1,label=$r_1$]{1}
\Vertex[x=2,y=1, label=$r_2$]{2}
\Vertex[x=3, label=$r_3$]{3}
\Vertex[x=2, y=-1, label=$r_4$]{4}
\Edge[Direct,label=2,position=above,bend=45](1)(2)
\Edge[Direct,label=2,position=above,bend=45](1)(4)
\Edge[Direct,label=1,position=above](2)(1)
\Edge[Direct,label=1,position=above,loopsize=0.5cm, loopposition=40,loopshape=45](2)(2)
\Edge[Direct,label=2,position=above](2)(3)
\Edge[Direct,label=2,position=above](2)(4)
\Edge[Direct,label=1,position=above,bend=-20](3)(2)
\Edge[Direct,label=1,position=above](4)(1)
\Edge[Direct,label=2,position=above, bend=-20](4)(2)
\end{tikzpicture}

\caption{An example to show how Algorithm~\ref{alg:WeiDNeR} works; taking the graph in the left, it produces the graph in the right.  }
\label{fig:weidner_example}
\end{figure}
\SetKwComment{Comment}{/* }{ */}
\begin{algorithm}
\caption{WeiDNeR - An algorithm to generate a directed and weighted relation-relation network} \label{alg:WeiDNeR}
\KwData{$T \gets Triples\ in\ KG$}
\KwResult{$N_{rel}$}
\For{each pair of relations \texttt{$<r_a, r_b>$}}{
\eIf{$r_a\neq r_b$}
{
    $\#direct_{<r_a, r_b>} \gets \big | \{(\langle h_1,r_a,t_1\rangle,\langle h_2,r_b,t_2\rangle) :\  \langle h_1,r_a,t_1\rangle \in T, \langle h_2,r_b,t_2\rangle \in T, t_1=h_2 \} \big |$; \label{alg:dir1}\\
    
    $\#direct_{\langle r_b, r_a\rangle} \gets \big | \{(\langle h_1,r_a,t_1\rangle,\langle h_2,r_b,t_2\rangle) :\  \langle h_1,r_a,t_1\rangle \in T, \langle h_2,r_b,t_2\rangle \in T, h_1=t_2 \} \big |$; \label{alg:dir2}
    
    $\#indirect \gets \big | \{(\langle h_1,r_a,t_1\rangle,\langle h_2,r_b,t_2\rangle) :\  \langle h_1,r_a,t_1\rangle \in T, \langle h_2,r_b,t_2\rangle \in T, (h_1=h_2 \lor t_1=t_2) \} \big |$;\label{alg:indir1}
     
    $Weight_{\langle r_a, r_b\rangle} = \#direct_{\langle r_a, r_b\rangle} + \#indirect$; \label{alg:sum1-starts}
    $Weight_{\langle r_b, r_a\rangle} = \#direct_{\langle r_b, r_a\rangle} + \#indirect$;
    
    \If{$Weight_{\langle r_a, r_b\rangle} > 0$}
    {
     $N_{rel}\gets N_{rel} \bigcup \{\langle r_a, r_b, Weight_{\langle r_a, r_b\rangle}\rangle$\};
    }
    \If{$Weight_{\langle r_b, r_a\rangle} > 0$}
    {
     $N_{rel}\gets N_{rel} \bigcup \{\langle r_b, r_a, Weight_{\langle r_a, r_b\rangle}\rangle \}$; 
    }\label{alg:sum1-ends}
}{
  
  $\#direct \gets \big | \{(\langle h_1,r_a,t_1\rangle,\langle h_2,r_b,t_2\rangle) :\ \langle h_1,r_a,t_1\rangle \in T, \langle h_2,r_b,t_2\rangle \in T, t_1=h_2, (h_1\neq t_1 \lor h_1\neq t_2) \} \big |$; \label{alg:dir3}
  
  $\#indirect \gets \big | \{(\langle h_1,r_a,t_1\rangle,\langle h_2,r_b,t_2\rangle) :\ \langle h_1,r_a,t_1\rangle \in T, \langle h_2,r_b,t_2\rangle \in T, ((h_1=h_2 \land t_1\neq t_2) \lor (t_1=t_2 \land  h_1\neq h_2)) \} \big |$;\label{alg:indir2}

  $Weight_{\langle r_a, r_a\rangle} = \#direct + \#indirect$; \label{alg:sum2-starts}
  
  \If{$Weight_{\langle r_a, r_a\rangle} > 0$}
  {  
  $N_{rel}\gets N_{rel} \bigcup \{r_a, r_a, Weight_{\langle r_a, r_a\rangle}\}$;
  }\label{alg:sum2-ends}
}
}
\end{algorithm}

\subsection{Node Embeddings}
Features for the nodes in a given network $N_{rel}$ could be generated leveraging the network's structural information.
In order to generate these features, 
in this work, Node2Vec~\cite{Grover2016Node2Vec} which learns continuous feature representations for nodes in networks with the likelihood of preserving neighborhood information is used. It applies second-order (biased) random walks to efficiently explore diverse neighborhoods of a given node and then makes use of the SkipGram~\cite{Mikolov2013SkipGram} word embedding method to learn embeddings by treating the generated walks as sentences. Given a network (such as $N_{rel}$), in order to select the next hop, Node2Vec computes second-order transition probabilities as shown in Equation~\ref{eq:node2vec}.

\begin{equation}
    p(u|v,t) = \frac{\alpha_{pq}(t,u)w(u,v)}{\sum_{u^{'} \in  N_{v}} \alpha_{pq}(t,u^{'})w(u^{'},v) }
    \label{eq:node2vec}
\end{equation}
where $u,v \in V$, $N_v$ denotes the neighboring nodes of $v$, and $w(u,v)$ is the weight of the edge between the nodes $u$ and $v$, and $\alpha$ is the bias factor used to reweigh the edge weights depending on the previous visited state and it is computed state as shown in Equation~\ref{eq:node2vec-alpha}.

\begin{equation}
    \alpha(t,v) = \begin{cases}
    \frac{1}{p},& \text{if  $d_{tv} = 0$}\\
    1,& \text{if $d_{tv}$ = 1 }\\
    \frac{1}{q},& \text{if $d_{tv} = 2$}
    \end{cases}
\label{eq:node2vec-alpha}
\end{equation}

where \textbf{$p$} is the \textit{return parameter} which controls the likelihood of immediately revisiting a node, $q$ is the \textit{in-out parameter} controlling the likelihood of revisiting a node’s one-hop neighborhood, and $d_{tv}$ is the shortest distance between the nodes $t$ and $v$.

These random walks are then passed to the Skip-gram model to 
learn the node embeddings. Since the Skip-gram model aims to learn continuous feature representations for words by optimizing a neighborhood preserving likelihood objective, in Node2Vec it could be interpreted as aiming to maximize the probability of predicting the correct context node $v$ for a given center node $u$. 


\subsection{Training Procedure}
The graph-based features for relations in training, validation, and test sets are created separately and used as inputs for when models are trained. 
In a truly-inductive setting, only the triples from the training set are used to generate the graph-based features for the relations appearing in the training. Similarly, for relations in the validation and test sets, only triples from the validation and test sets are used respectively. This is performed in order to avoid using information from the unseen graphs (i.e., from validation and test sets) to learn features during training. Similarly, in both semi-inductive and fully-inductive settings, only the triples from the training set are taken as input to generate the graph-based feature for the relations.

Once the features of the entities and relations are generated or encoded, then they are used to optimize the model for LP by applying stochastic gradient descent. For each positive triple $<e_i, r_j, e_k>$, a positive score $S_p$ is computed. Then, a corrupted negative triple is created by replacing the head or the tail entity with a random entity, and its score $S_n$ is computed.

\subsection{Computational complexity}
As it is discussed in the previous sections, RAILD uses text-based and graph-based encoders. The text-based encoder is used for both entities and relations whereas the graph-based encoder is used only to encode relations. Note that the graph-based features for relations are pre-computed and hence, the major part of the computational cost of training the model comes from text-based encoder.  The BERT encoder used has a complexity of \texttt{$\bigo(n^2)$} for encoding a sentence of length n. This entails that for training RAILD, the time complexity would be \texttt{$\bigo(\left|T\right|n^2)$} where \texttt{T} is the set of triples. The length of sentences n is in practice fixed and assuming the n is the same for all entities and relations, the complexity would remain linear with respect to the number of triples in the KG, up to a constant factor. 

During testing, the text-based encoder is applied only for unseen entities and unseen relations while the embeddings for seen entities and seen relations can be pre-computed. Hence, the LP for a given entity and relation is linear in the number of entities and relations in the graph.



%% file: tex/Experiments.tex
\section{Experiments}\label{sec:experiments}
In this section, the details on experimentation including the baselines, the datasets, the experimentation settings, and the results are discussed. Our implementation and the datasets are made publicly available\footnote{https://github.com/GenetAsefa/RAILD}. 

\subsection{Datasets}\label{sec:new-dataset}
The three inductive LP settings discussed in Section~\ref{sec:related-work} 
are considered for experimentation. In order to evaluate RAILD in the semi-inductive setting the datasets FB15K-237~\cite{toutanova-chen-2015-observed} and WN18RR~\cite{dettmers2018convolutional} with the splits provided in~\cite{Daza2021InductiveER}, are used. In a fully-inductive setting, the model is evaluated on dataset Wikidata5M~\cite{Wang2021KEPLERAU} and compared against SOTA models. The statistics of these datasets are provided in Table~\ref{tab:datasets}. To the best of our knowledge, there are no benchmark datasets that contain unseen relations in their validation and test sets. To address this issue and to enable the evaluation of RAILD with unseen relations, a new dataset Wikidata68K is created taking Wikidata5M as raw data. The pipeline developed to create this dataset is inspired by~\cite{gesese2021LitWD} and is constituted of the following steps.   

\begin{enumerate}
    \item Input: 
            Raw data $T$ containing triples from which the dataset will be created, and a set of pairs of relations and their types $RT$. In Wikidata, there exists a metaclass (\texttt{Q107649491}: type of Wikidata property) with instances that are types of properties (i.e., relations). For example, \texttt{Q29546443} (Wikidata property for items about books) is an instance of \texttt{Q107649491} and the property \texttt{P123(publisher)} is an instance of \texttt{Q29546443}. Therefore, \texttt{(P123, Q29546443)} could be an entry in $RT$. 
    \item Removing relations which occur in less than $N$ number of triples ($N=3$, for Wikidata68K).
    \item Removing inverse relations, entities and relations without a label, and duplicate relations.
    \item Randomly splitting the set of relations into three $R_1$, $R_2$, and $R_3$ while trying to keep the same type of relations in the same set based on $RN$ and extract their corresponding triples $T_1$, $T_2$, and $T_3$ from $T$.
    \item Creating K-cores for each of $T_1$, $T_2$, and $T_3$. (For Wikidata68K, the value of $k$ is set to 10, 6, and 5 for $T_1$, $T_2$, and $T_3$ respectively).
    \item Removing relations which are skewed towards either the head or the tail at least 50\% of the time, from each of $T_1$, $T_2$, and $T_3$.
\end{enumerate}

\begin{table}[]
    \centering
    \caption{Dataset statistics}
    \resizebox{9cm}{!}{
    \begin{tabular}{ccccc}
    \toprule
         & \textbf{WN18RR} & \textbf{FB15K-237} & \textbf{Wikidata5M} & \textbf{WD20K(25)}\\
    \midrule
       Relations & 11 & 237 & 822 & 333 \\
       \hline
      & \multicolumn{4}{c}{Training} \\
       \cline{2-5}
       Entities & 32,755 & 11,633 & 4,579,609 &17,275\\
       Triples & 69,585 & 215,082 & 20,496,514 & 38,023 \\
      \hline
      & \multicolumn{4}{c}{Validation} \\
       \cline{2-5}      
       Entities & 4,094 & 1,454 & 7,374 & 3,092\\
       Triples & 11,381 & 42,164 & 6,699 & 4,072\\
       \hline
      & \multicolumn{4}{c}{Test} \\
       \cline{2-5}
       Entities & 4,094 & 1,454 & 7,475 & 2,615 \\
       Triples & 12,087 & 52,870 & 6,894 & 3,329 \\
       \midrule
       \multicolumn{4}{c}{\textbf{Wikidata68K}} & \\
       \cline{1-4}
         & Training & Validation & Test &\\
      \cline{1-4}
       Entities  & 55,488  & 6,559 & 5,813&\\
       Relations  & 72  & 37 & 44&\\
      Triples & 667,413 & 67,892 & 45,512& \\
       \cline{1-4}
    \end{tabular}
    }
    \label{tab:datasets}
\end{table}
\subsection{Baselines}\label{baselines}
For semi-inductive and fully-inductive settings, RAILD could be compared with SOTA models like BLP and KEPLER on FB15K-237, WN18RR, and Wikidata5M datasets. 
Since there exists no SOTA model which handles unseen relations, four different baselines Glove-BOW$_t$, Glove-DKRL$_t$, BE-BOW$_t$, and BE-DKRL$_t$ are created by extending the baselines in BLP, i.e., Glove-BOW, Glove-DKRL, BE-BOW, and BE-DKRL respectively to also encode relations using their textual descriptions in the same way they encode entities. These models are different re-implementations of DKRL~\cite{Xie2016DKRL} where Glove-DKRL uses Glove embeddings as an input to the DKRL architecture whereas Glove-BOW is the Bag-Of-Word baseline of DKRL. Furthermore,  BE-BOW, and BE-DKRL are other varieties that use context-insensitive BERT Embeddings (BE). Note that the baselines created in this work are used for evaluation in the truly-inductive setting on Wikidata68K dataset and to compare them with RAILD.

\subsection{Experimentation Setting}

\noindent{\textbf{Scoring.}} TransE, SimplE, DistMult, and ComplEx are some of the well known translational models with \texttt{TransE} being the simplest among all. 
ComplEx handles antisymmetric relations better than both TransE and DistMult~\cite{Trouillon2016ComplEx}. However, TransE could also perform well in some cases, for example, in BLP the best performing scoring function is TransE followed by ComplEx. Hence, in this paper, the scoring functions \texttt{TransE} and \texttt{ComplEx} are  selected. 

\noindent{\textbf{Model Selection}} For the Node2Vec model, number of walks=1000, length=10, window size=10, epochs=100, dim=768 are used for FB15K-237 with semi-inductive split and Wikidata5M with fully-inductive split. For WN18RR with semi-inductive split, number of walks=5000, length=10, window size=5, epochs=100, dim=768 are used. For Wikidata68K, number of walks=10, length=10, window size=10, epochs=100, and dim=768.

Similar to~\cite{Daza2021InductiveER}, for all newly created baselines and RAILD models, a grid search is run on FB15K-237 and the hyperparameter values with the best performance on the validation set are chosen. Then, these values are reused for training with the other datasets. 
For the BOW and DKRL baselines, inspired by~\cite{Daza2021InductiveER}, learning rate: {1e-5, 1e-4, 1e-3}, L2 regularization coefficient: {0, 1e-2, 1e-3} are applied. Adam optimizer is used with no learning rate schedule, and the models are trained for 80 epochs with a batch size of 64 with WN18RR, FB15k-237, and 40 epochs with a batch size of 254 with Wikidata68K.

For the RAILD models, loss function: {margin, negative log-likelihood}, learning rate: {1e-5, 2e-5, 5e-5}, L2 regularization coefficient: {0, 1e-2, 1e-3} are used. Adam optimizer with a learning rate decay schedule with a warm-up for 20\% of the total number of iterations is used. The models are trained for 40 epochs (80 epochs for models which combine text and graph-based features for relations) with a batch size of 64 with WN18RR and FB15k-237, and 5 epochs with a batch size of 128 with Wikidata5M. In all the experiments, the negative sample size is set to 64.

\subsection{Results}
Two main varieties of RAILD, i.e., {\tt RAILD-TransE} and \texttt{RAILD-ComplEx}, are created with the scoring functions TransE and ComplEx respectively.  
The results obtained with the different inductive LP settings are discussed in the subsequent sections.

\subsubsection{\textbf{Results in semi-inductive setting}}
The results obtained with the semi-inductive setting on 
WN18RR and FB15K-237 are shown in Table~\ref{tab:res-semi}. 
The table compares 2 different varieties of RAILD (i.e., RAILD-TransE and RAILD-ComplEx) 
with the different models from~\cite{Daza2021InductiveER} and our baselines.
These results show that \texttt{RAILD-TransE} outperforms all the other models on FB15K-237 w.r.t. all metrics. On the contrary, on WN18RR RAILD-ComplEx provides the best result w.r.t. all metrics whereas the second best results are obtained with RAILD-TransE w.r.t. all metrics except Hits@1. 

Although TransE is a less elaborate 
model than ComplEx, it provides better results 
when 
used with RAILD on FB15K-237 and competitive results on WN18RR. Same is the case with the results obtained in the truly-inductive setting on Wikidata68K (see Section~\ref{t-ind-results}). This suggests that the expressiveness of TransE could be highly improved 
with RAILD which has a more expressive encoder.

\begin{table*}[]
    \centering
    \caption{LP results on \textbf{semi-inductive} setting on WN18RR and FB15K-237 datasets}

    \begin{tabular}{ccccccp{0.5cm}cccc}
         \toprule
         & & \multicolumn{4}{c}{FB15K-237} & &\multicolumn{4}{c}{WN18RR} \\
         \cline{3-6} \cline{8-11} 
         & & MRR & Hits@1 & Hits@3 & Hits@10 & & MRR & Hits@1 & Hits@3 & Hits@10  \\
         \hline
        \multirow{8}{*}{\shortstack[l]{S/F-inductive\\models}}&Glove-BOW$^{*}$ & 0.172& 0.099 & 0.188 & 0.316 & & 0.170 & 0.055 & 0.215 & 0.405 \\
        
        & Glove-DKRL$^{*}$ & 0.112 & 0.062 & 0.111 & 0.211 & & 0.115 & 0.031 & 0.141 & 0.282 \\ 
        & BE-BOW$^{*}$  & 0.173 & 0.103 & 0.184 & 0.316 & & 0.180 & 0.045 & 0.244 & 0.450 \\
         & BE-DKRL$^{*}$ & 0.144 & 0.084 & 0.151 & 0.263 & & 0.139 & 0.048 & 0.169 & 0.320 \\
          & BLP-TransE$^{*}$ & 0.195 & 0.113 & 0.213 & 0.363 & & 0.285 & 0.135 & 0.361 & 0.580 \\
        & BLP-DistMult$^{*}$ &  0.146 & 0.076 & 0.156 & 0.286 & & 0.248 & 0.135 & 0.288 & 0.481 \\
        & BLP-ComplEx$^{*}$ &  0.148 & 0.081 & 0.154 & 0.283 & & 0.261 & 0.156 & 0.297 & 0.472\\
        & BLP-SimplE$^{*}$ & 0.144 & 0.077 & 0.152 & 0.274 & & 0.239 & 0.144 & 0.265 & 0.435  \\
        \hline
        \multirow{6}{*}{\shortstack[l]{\textbf{Ours}}}&Glove-BOW$_{t}$ & 0.1464 & 0.0813 & 0.1636 & 0.2681 & & 0.1589 & 0.0465 & 0.2085 & 0.3812\\   
        & Glove-DKRL$_{t}$ &  0.1131 & 0.0678 & 0.1176 & 0.1990 & & 0.1111 & 0.0283 & 0.1362 & 0.2749 \\
        & BE-BOW$_{t}$ &  0.1569 & 0.0857 & 0.1780 & 0.2923 & & 0.1810 & 0.0424 & 0.2483 & 0.4529 \\
        & BE-DKRL$_{t}$  & 0.1385 & 0.0817 & 0.1473 & 0.2477 & & 0.1342 & 0.0461 & 0.1636 & 0.3090 \\
        \cline{2-11}

        & RAILD-TransE & \textbf{0.2163} & \textbf{0.1268} & \textbf{0.2411} & \textbf{0.3974} & & 0.2909 & 0.1360 & 0.3689 & 0.5997  \\
        & RAILD-ComplEx & 0.1971 & 0.1169 & 0.2121 & 0.3639 & & \textbf{0.3204} & \textbf{0.1772} & \textbf{0.3895} & \textbf{0.6087} \\
         \bottomrule
    \end{tabular}

    \label{tab:res-semi}
\end{table*}
\begin{table*}[]
    \centering
    \caption{Ablation studies with all 4 datasets using TransE scoring function. }
 
    \begin{tabular}{ccccccp{0.5cm}cccc}
         \toprule
         & & \multicolumn{4}{c}{FB15K-237} & &\multicolumn{4}{c}{WN18RR} \\
         \cline{3-6} \cline{8-11} 
         & & MRR & Hits@1 & Hits@3 & Hits@10 & & MRR & Hits@1 & Hits@3 & Hits@10  \\
         \hline
        & RAILD-TransE & \textbf{0.2163} & \textbf{0.1268} & \textbf{0.2411} & \textbf{0.3974} & & \textbf{0.2909} & 0.1360 & \textbf{0.3689} & \textbf{0.5997}  \\
        & RAILD-TransE(w/o feat)  & 0.2130 & 0.1267 & 0.2363 & 0.3872 & & 0.2906 & \textbf{0.1377} & 0.3672 & 0.5944\\ 
        & RAILD-TransE(w/o txt) & 0.2030 & 0.1168 & 0.2258 & 0.3777 & & 0.2855 & 0.1312 & 0.3640 & 0.5945 \\
        & & \multicolumn{4}{c}{Wikidata68K} & &\multicolumn{4}{c}{Wikidata5M} \\
        \cline{3-6} \cline{8-11} 
        & RAILD-TransE & \underline{0.0285} & \underline{0.0059} & \textbf{0.0283} & \textbf{0.0688} & & \textbf{0.4551} & 0.2200 & \textbf{0.6345} & \textbf{0.8489} \\
        & RAILD-TransE(w/o feat) &\textbf{0.0300} & \textbf{0.0077} & \textbf{0.0283} & 0.0661 & & 0.4529 & 0.2274 & 0.6190 & 0.8376\\
        & RAILD-TransE(w/o txt) & 0.0137 & 0.0014 & 0.0130 & 0.0320 && 0.4522 & \textbf{0.2304} & 0.6163 & 0.8378\\
         \bottomrule
    \end{tabular}

    \label{tab:ablation}
\end{table*}

\subsubsection{\textbf{Results in fully-inductive setting}}
Table~\ref{tab:res-wikidata5m} shows the results obtained with the fully-inductive setting on the dataset Wikidata5M. 
Due to limited computational resources, for the experiment on Wikidata5M the distilled version of Bert, i.e., DistilBert~\cite{corr/abs-1910-01108} is used since it is cheaper to train as compared to Bert. However, it should be noted that since DistilBert is a slimmed-down version of BERT with fewer parameters it may lead it to be less powerful than Bert. 
Although MLMLM is not an embedding-based LP model, it is compared with our approach on Wikidata5M. 
It can be seen that even with DistilBert RAILD-TransE trained on Wikidata5M outperforms KEPLER and MLMLM w.r.t. almost all metrics.


\begin{table}[]
    \centering
    \caption{LP results on Wikidata5M dataset using DistilBERT instead of BERT for RAILD models}
    \label{tab:res-wikidata5m}
 
    \resizebox{8cm}{!}{
    \begin{tabular}{p{2cm}cccc}
    \toprule
    & MRR & Hits@1 & Hits@3 & Hits@10 \\
    \midrule
    Glove-BOW$^{*}$& 0.343 &0.092 &0.531 &0.756\\
    Glove-DKRL$^{*}$ &0.362 &0.082 &0.586 &0.798\\
    BE-BOW$^{*}$ & 0.282& 0.077 &0.403 &0.660\\
    BE-DKRL$^{*}$ &0.322 &0.097 &0.474 &0.720\\
    BLP-TransE$^{*}$ & \textbf{0.478} & \textbf{0.241} & \textbf{0.660} & \textbf{0.871}\\
    KEPLER~\cite{Wang2021KEPLERAU}  &0.402 &0.222& 0.514 &0.730 \\
    MLMLM~\cite{Clouatre2021MLMLM} & 0.284 & 0.226 & 0.285 & 0.348\\
    RAILD-TransE (\textbf{DistilBERT})& \underline{0.4551} & 0.2200 & \underline{0.6345} & \underline{0.8489} \\
    \bottomrule
    \end{tabular}
    }
\end{table}

\subsubsection{\textbf{Results in truly-inductive setting}}\label{t-ind-results}
Table~\ref{tab:res-Wikidata68K} presents the results obtained on Wikidata68K (see Section~\ref{sec:new-dataset} for details).
The set of training, validation, and test relations are mutually exclusive.
Moreover, 89\% of validation entities and 74\% of test entities are not seen during training. For datasets like Wikidata68K, it is not possible to just randomly initialize the relations (i.e., it is required for an LP model to have features for relations). Therefore, in order to assess the capability of the proposed model RAILD on such a challenging dataset (Wikidata68K), the baselines discussed in Section~\ref{baselines} are created. The best results on this dataset are obtained with RAILD-TransE w.r.t. all the metrics.  

As compared to the other datasets, the results obtained on Wikidata68K, in general, are low. This is mostly attributed to the nature of the dataset as explained above, i.e., the relations sets in the train validation and test sets being 100\% mutually exclusive. Moreover, the WeiDNeR algorithm is applied to the training set, the validation set, and the test set separately so as to avoid generating features using unseen graphs for training. As this is the first work, to the best of our knowledge, to ever make an attempt to perform LP with unseen relations, it would facilitate further research in the community to redirect the focus to unseen relations as well as entities.

\begin{table}[]
    \centering
    \caption{LP results on Wikidata68K datasets}
    \label{tab:res-Wikidata68K}

     \resizebox{.45\textwidth}{!}{
    \begin{tabular}{ccccc}
    \toprule
    & MRR & Hits@1 & Hits@3 & Hits@10 \\
    \midrule
    Glove-BOW$_{t}$ & 0.0119 & 0.0005 & 0.0146 & 0.0295  \\
    Glove-DKRL$_{t}$ & 0.0031 & 0.0005 & 0.0029  & 0.0064\\
    BE-BOW$_{t}$ & 0.0184 & 0.0005 & 0.0225 & 0.0474 \\
    BE-DKRL$_{t}$ & 0.0055 & 0.0006 & 0.0052 & 0.0125 \\
    \hline
    RAILD-TransE &  \textbf{0.0285} & \textbf{0.0059} & \textbf{0.0283} & \textbf{0.0688} \\
    RAILD-ComplEx &  0.0157 & 0.0027 & 0.0125 & 0.0351 \\
    \bottomrule
    \end{tabular}
    }
\end{table}


\subsubsection{\textbf{Ablation studies}}
As the results given in Table~\ref{tab:ablation} for the datasets FB15K-237, WN18RR, and Wikidata5M indicate,  according to almost all the metrics, combining text-based and graph-based features for relations (i.e., RAILD-TransE) provides better results than using them separately. Moreover, RAILD-TransE(w/o text) model variant which uses only graph-based relation features is competitive with its counterpart text-based variant RAILD-TransE(w/o feat) and  specially on Wikidata5M it even provides slightly better hits@1 and hits@10 results than RAILD-TransE(w/o feat).    This indicates that the RAILD-TransE(w/o text) could be used in cases where KGs do not contain textual descriptions for their relations. Similarly, combining the two kind of features for Wikidata68K provides better results w.r.t. hits@10 and equal or competitive results w.r.t. the other metrics as compared to RAILD-TransE(w/o feat).     

Note that both RAILD-TransE(w/o feat) and RAILD-TransE(w/o txt) outperform the BLP models which share the same scoring function. For instance, RAILD-TransE(w/o txt) which uses only graph-based features for relations outperforms all the baselines including BLP-TransE~\cite{Daza2021InductiveER} which randomly initializes relations, on both datasets FB15K-237 and WN18RR w.r.t. almost all the metrics.

\subsubsection{\textbf{Additional Experiments}}
In addition to the experiments discussed above 
further experiments are also performed to compare the performance of the proposed model with QBLP~\cite{Ali2021ImprovingIL} which is an inductive LP model developed for hyper-relational KG. The same set of optimal hyperparameter values from FB15K-237 datasets are used.
RAILD is compared with QBLP on a WD20K(25) dataset~\cite{Ali2021ImprovingIL} 
for 
hyper-relational KG.  The dataset statistics considering only the triples (removing qualifiers) are given in Table~\ref{tab:res-semi} and the results in Table~\ref{tab:res-qblp}. 
The performance of our model could be negatively impacted by the fact that the size of this dataset is very small as compared to the other datasets used in this work such as FB15K-237. Moreover, since there are relations that occur only in a few triples, generating relations features using WeiDNeR could not be applied as these relations become outliers, i.e., they could not be linked to any other relation (see Algorithm~\ref{alg:WeiDNeR}). 
Therefore, only RAILD-TransE(w/o feat) could be used. Despite the argument stated above, RAILD scores the best Hits@10 as compared to QBLP which makes use of 
qualifiers. 

\begin{table}[]
    \centering
    \caption{LP results with semi-inductive setting on \textbf{WD20K(25)} dataset. \#QP denotes the number of qualifiers per statement.}

    \resizebox{8cm}{!}{
    \begin{tabular}{ccccc}
    \toprule
        & \#QP & MRR & Hits@1  & Hits@10 \\
       \hline
       BLP-TransE$^*$ & 0 & 0.1245 & 0.0598  & 0.2343   \\
       QBLP$^*$ & 0 & 0.1702 & 0.0882 & 0.2950 \\
       QBLP$^*$ & 2 & 0.2036 & 0.1177 & 0.3226 \\
       QBLP$^*$ & 4 & \textbf{0.2105} & \textbf{0.1232} & 0.3009 \\
       QBLP$^*$ & 6 & 0.1950 & 0.1114 & 0.3160 \\
       RAILD-TransE (w/o feat) & 0 & 0.1586& 0.0761 & \textbf{0.3313} \\
    \bottomrule
    \end{tabular}
    }
    \label{tab:res-qblp}
\end{table}

%% file: tex/Conclusion.tex
\section{Conclusion and Future Work}\label{sec:conclusion}
In this work, a novel inductive LP model \texttt{RAILD} which handles unseen relations is introduced. It works by fine-tuning pretrained LMs with an LP objective. Textual descriptions of entities and relations are used to generate features for the corresponding entities and relations. Moreover, a novel algorithm, i.e., WeiDNeR, is proposed to generate a directed and weighted relation-relation network given a KG. 
The results of the extensive experiments indicate that using graph structure information to generate relations brings an improvement over models which randomly initialize relations. Moreover, for relations with textual descriptions, it is possible to use the structured information to encode relations and hence, learn embeddings for unseen relations. As a future direction, the proposed model will be adapted to hyper-relational KGs. Moreover, the WeiDNeR algorithm will be further investigated for the LP task where few-shot relations exist.